\documentclass[letterpaper]{article} 
\usepackage[preprint]{2027}  
\usepackage[hyphens]{url}  
\usepackage{graphicx} 
\usepackage{natbib}  
\usepackage{caption} 
\usepackage{algorithm}
\usepackage{algorithmic}

\usepackage{newfloat}
\usepackage{listings}
\DeclareCaptionStyle{ruled}{labelfont=normalfont,labelsep=colon,strut=off} 
\floatstyle{ruled}
\newfloat{listing}{tb}{lst}{}
\floatname{listing}{Listing}

\usepackage{booktabs}

\usepackage{amsmath}
\usepackage{amssymb}
\usepackage{booktabs}        
\usepackage{multirow}        
\usepackage{graphicx}        
\usepackage{threeparttable}  
\usepackage{natbib}          
\usepackage[table]{xcolor}
\usepackage{subcaption}
\usepackage{enumitem}
\usepackage{bm}

\definecolor{improvegreen}{HTML}{00B050}

\usepackage{makecell}

\usepackage{xcolor}

\usepackage{hyperref}

\title{Continue or Replan? Bernoulli-Continuation Policy Learning \\ for Adaptive Horizon Execution}
\author {
    Weichen Xu\textsuperscript{\rm 1,\rm 2,\(\ddagger\)}\equalcontrib,
    Zhenhua Liu\textsuperscript{\rm 2,}\equalcontrib,
    Lin Luo\textsuperscript{\rm 2},
    Yaobo Liang\textsuperscript{\rm 2},
    Chengtang Yao\textsuperscript{\rm 2},
    Qingyu Mei\textsuperscript{\rm 1,\rm 2,\(\ddagger\)},
    Jian Cao\textsuperscript{\rm 1},
    Xixin Cao\textsuperscript{\rm 1},
    Xing Zhang\textsuperscript{\rm 1},
    Jiaolong Yang\textsuperscript{\rm 2,}\corresponding,
    Baining Guo\textsuperscript{\rm 2}
}
\affiliations {
    \textsuperscript{\rm 1}Peking University\\
    \textsuperscript{\rm 2}Microsoft Research Asia\\
    xuweichen1999@stu.pku.edu.cn, zhenhualiu@microsoft.com
 , jiaoyan@microsoft.com
}

\begin{document}

\maketitle

\begingroup
\renewcommand{\thefootnote}{\(\ddagger\)}
\footnotetext[0]{Work done during an internship at Microsoft Research Asia.}
\endgroup

\begin{abstract}

Existing chunk-based Vision-Language-Action (VLA) models execute a fixed number of actions (i.e., execution horizon) before replanning, turning replanning into a task-agnostic periodic schedule that is independent of task progress.
As a result, when no replanning boundary falls before a critical manipulation stage, it is executed from a stale chunk rather than a freshly replanned one.
To address this limitation, we propose \textbf{B}ernoulli-\textbf{C}ontinuation \textbf{P}olicy (\textbf{BCP}), a lightweight, plug-and-play framework for adaptive horizon execution that keeps the base VLA frozen. 
Given a fixed-length action chunk, its continuation head decomposes execution-horizon selection into a sequence of continue-or-replan decisions, which imposes an ordinal, prefix-sharing inductive bias over candidate horizons rather than treating them as independent classes.
Since the optimal horizon for each chunk is not observable, we train this head with reinforcement learning from trajectory-level outcomes and introduce a Replanning-Efficiency Reward that jointly rewards task success and efficient VLA usage, discouraging the policy from collapsing to unnecessarily short horizons. 
On RoboTwin 2.0 with LingBot-VLA as the base policy, BCP improves the average success rate by \textbf{+11.08\%} on 13 low-success tasks and from 89.88\% to \textbf{93.94\%} (\textbf{+4.06\%}) across all 50 tasks.
Although trained only under the Clean setting, BCP generalizes to the Randomized setting, raising the average success rate by \textbf{+4.06\%}. It also transfers to a different base policy $\pi_{0.5}$, achieving a better result on LIBERO (+1.7\%) and, notably, on the harder LIBERO-PRO (\textbf{+6.8\%}), where its advantage widens as tasks become more difficult. On a real robot, BCP lifts success from 74\% to 92\% and from 44\% to 84\% on two manipulation tasks. Meanwhile, its negligible overhead, combined with higher success, makes BCP's overall runtime even lower than the fixed-horizon baselines. 
More details can be found at \href{https://fleetfootwork.github.io/BCP/}{Project Page}.

\end{abstract}

\section{Introduction}

Vision-Language-Action (VLA) models have recently shown strong potential for language-conditioned robotic manipulation. A common design in recent VLA systems is action chunking~\cite{zhao2023learning}, where the model predicts a sequence of future actions rather than a single next action. This design reduces the frequency of expensive model inference and improves temporal consistency~\cite{chi2025diffusion}. In a chunk-based VLA, the prediction horizon determines how many future actions the model generates, while the execution horizon determines how many of these predicted actions are actually executed before the robot observes the environment again and replans.

\begin{figure}[t]
\centering

\begin{minipage}[c]{0.99\linewidth}
\centering
\includegraphics[width=1.0\textwidth]{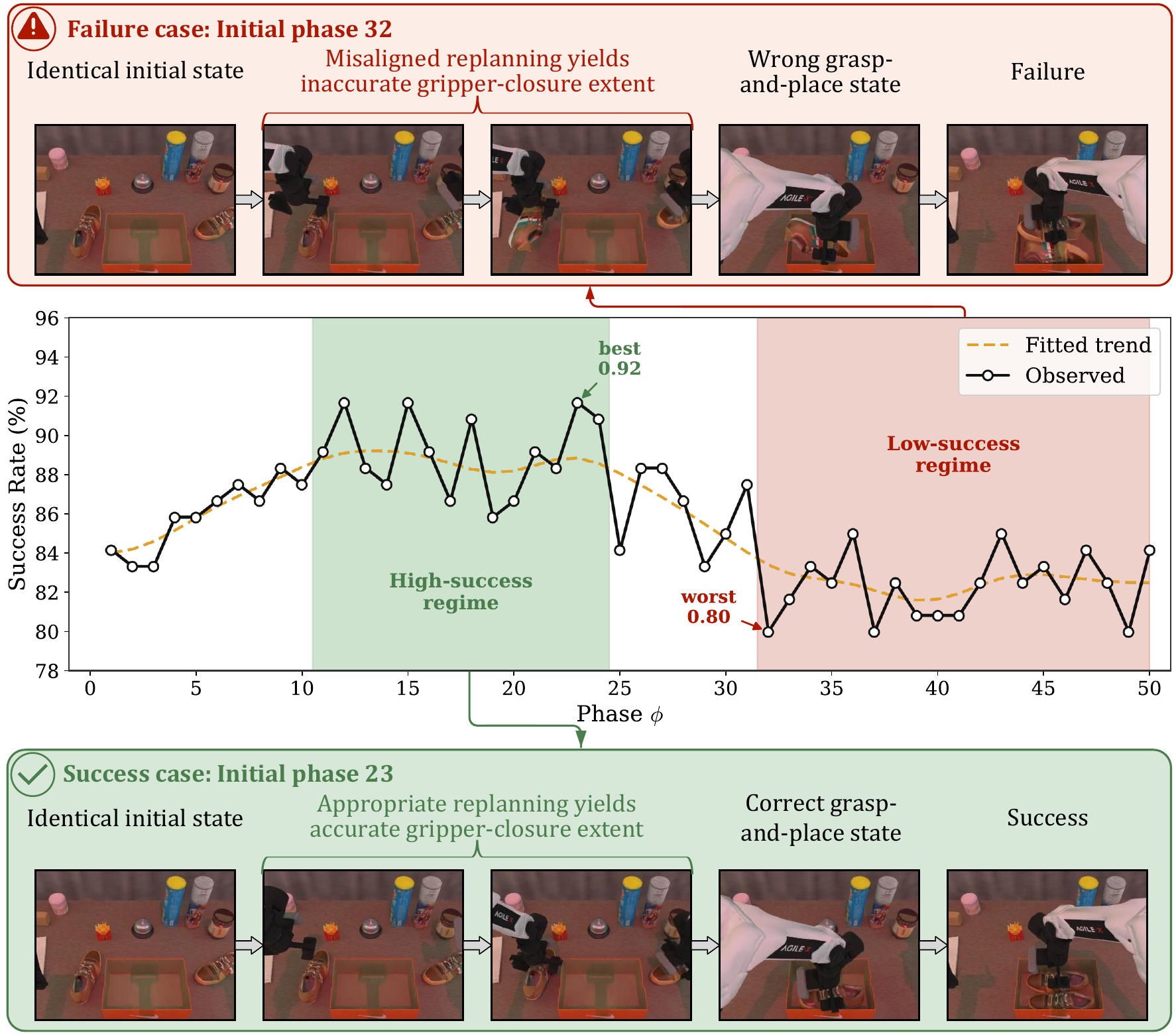}
\end{minipage}

\setlength{\abovecaptionskip}{4pt}
\setlength{\belowcaptionskip}{-16pt}
\caption{Phase-shift experiment with LingBot-VLA on RoboTwin 2.0 \emph{Placing Dual Shoes} task, sweeping all 50 initial phases $\phi$ under the same fixed 50-step horizon. Success rate varies strongly with $\phi$ (best $0.92$, worst $0.80$). At $\phi=32$ the robot executes an outdated gripper-closure extent and fails, while at $\phi=23$ a replanning boundary falls before the manipulation stage, yielding an accurate grasp and success.}
\label{fig:1}
\end{figure}

Existing chunk-based VLA models~\cite{kim2024openvla,black2024pi_0,wu2026pragmatic} adopt a fixed execution horizon throughout an episode. Though simple, this turns replanning into a periodic schedule independent of task progress. During coarse or free-space motion, executing a longer action sequence in a previously predicted chunk can provide smooth actions robust for many steps. At precision-critical stages such as manipulation, small pose or contact errors can quickly accumulate. The robot must therefore re-observe and replan prior to entering these stages, rather than relying on an earlier observation. A fixed schedule cannot guarantee this: when no replanning boundary falls before a critical stage, the robot is forced to execute it from a stale chunk.

\begin{table}[t]
\centering
\setlength{\abovecaptionskip}{4pt}
\setlength{\belowcaptionskip}{-16pt}
\setlength{\tabcolsep}{3pt}
\begin{tabular}{lcccc}
\toprule
 & \textbf{Original} & \textbf{Upper Bound} & \textbf{Lower Bound} & \textbf{Gap} \\
\midrule
\textbf{SR} & 88.52\% & 93.65\% & 82.50\% & \textbf{\textcolor{red}{11.30\%}} \\
\bottomrule
\end{tabular}
\caption{Phase-shift performance of LingBot-VLA on 50 RoboTwin 2.0 tasks. Upper and Lower Bounds take the best- and worst-performing phase \emph{per task}, so they are per-task oracles attainable only in hindsight. \textbf{SR} denotes Success Rate.}
\label{tab:phase_shift}
\end{table}

We demonstrate this limitation in a simple phase-shift experiment. All variants run the same fixed 50-step horizon but start from different phase $\phi$: the robot executes only $\phi$ actions in the first chunk and then replans every 50 steps, so they differ solely in \emph{when} their replanning boundaries fall.
As shown in Table~\ref{tab:phase_shift}, this single timing choice moves the success rate over an \textbf{11.30\%} gap on 50 RoboTwin 2.0~\cite{chen2025robotwin} tasks. 
Even this per-task upper bound is achievable only with hindsight. Critical moments occur at different times across tasks and stages, so no single phase can align replanning with all of them.
Figure~\ref{fig:1} illustrates this on \emph{Place Dual Shoes} task: at $\phi=32$, the robot commits an inaccurate gripper-closure and fails, whereas at $\phi=23$ a boundary falls before the manipulation stage and it succeeds. Success depends not on horizon length alone, but also on whether a boundary lands before each critical stage, 
which a fixed schedule cannot guarantee.

Since the desired replanning boundary varies along the trajectory, the execution horizon should not be a fixed hyperparameter but selected per chunk, so the robot re-observes before precision-sensitive stages. The key question is: \textbf{given a predicted action chunk, how long should the robot execute it before stopping and replanning?}

Learning such a policy is non-trivial and raises three challenges. 
First, candidate horizons have an inherent ordinal and prefix-sharing structure because a longer horizon subsumes shorter one. A standard softmax classifier treats them as unordered and independent classes, discarding this structure.
Second, per-chunk supervision is not identifiable because each horizon decision is coupled with subsequent chunks and later decisions. As the same outcome can arise from different horizon sequences, no unique ground-truth stopping label exists.
Third, even with trajectory-level rewards, a success-only signal drives the policy towards overly short horizons, causing excessive replanning and wasted VLA calls.

To address these challenges, we propose the \textbf{B}ernoulli-\textbf{C}ontinuation \textbf{P}olicy (\textbf{BCP}), a lightweight, plug-and-play framework that keeps the base VLA frozen. BCP decomposes execution-horizon selection into a sequence of continue-or-replan decisions: a longer horizon is selected only when the policy repeatedly decides to continue. This imposes an ordinal, prefix-sharing inductive bias over candidate horizons, addressing the first challenge.
For the second, we train BCP with reinforcement learning (RL) on trajectory-level outcomes rather than per-chunk stopping labels, naturally capturing the coupled effects of consecutive decisions. 
To resolve the success-efficiency trade-off, a Replanning-Efficiency Reward is introduced that jointly accounts for task success and VLA cost, favoring successful trajectories with fewer replanning steps and preventing the policy from collapsing to unnecessarily short horizons. 
With these three components, the execution horizon is no longer a fixed hyperparameter but a per-chunk decision learned from task outcomes, placing replanning boundaries where they matter.

We evaluate BCP on RoboTwin 2.0~\cite{chen2025robotwin} with ACT~\cite{zhao2023learning}, ABot-M0~\cite{yang2026abot} and LingBot-VLA~\cite{wu2026pragmatic}. Take LingBot-VLA as an example, BCP improves the success rate by \textbf{11.08\%} on 13 low-success tasks and raises the 50-task average from 89.88\% to \textbf{93.94\%} (\textbf{+4.06\%}), reaching state-of-the-art for VLAs. Although trained only on the Clean setting, BCP still generalizes to the Randomized setting (88.78\% to \textbf{92.84\%}, \textbf{+4.06\%}). With a different base policy $\pi_{0.5}$, it transfers to LIBERO and the harder LIBERO-PRO, the success rates are improved by +1.7\% and \textbf{+6.8\%}. On a real AGIBOT G1 robot, BCP improves success from 74\% to 92\% on \emph{grasping bottle} task and from 44\% to 84\% on \emph{hanging mug} task.

Our contributions are summarized as follows:
\begin{itemize}[leftmargin=15pt, itemsep= 0pt]
\item We identify replanning-timing misalignment as a key failure mode of fixed-horizon chunk execution, and formulate adaptive execution-horizon selection as a sequence of continue-or-replan decisions that captures the ordinal, prefix-sharing structure of candidate horizons.
\item We propose the Bernoulli-Continuation Policy, a lightweight, plug-and-play framework that keeps the base VLA frozen and is trained with reinforcement learning from trajectory-level outcomes.
\item We introduce a Replanning-Efficiency Reward that jointly encourages task success and execution efficiency, preventing the policy from collapsing to unnecessarily short horizons.
\item We demonstrate on RoboTwin 2.0, LIBERO, and real robot that BCP achieves better performance than baselines, generalizes to Randomized settings, and preserves the runtime efficiency of action chunking.
\end{itemize}

\begin{figure*}[htbp]
\centering

\begin{minipage}[c]{0.98\linewidth}
\centering
\includegraphics[width=1.0\textwidth]{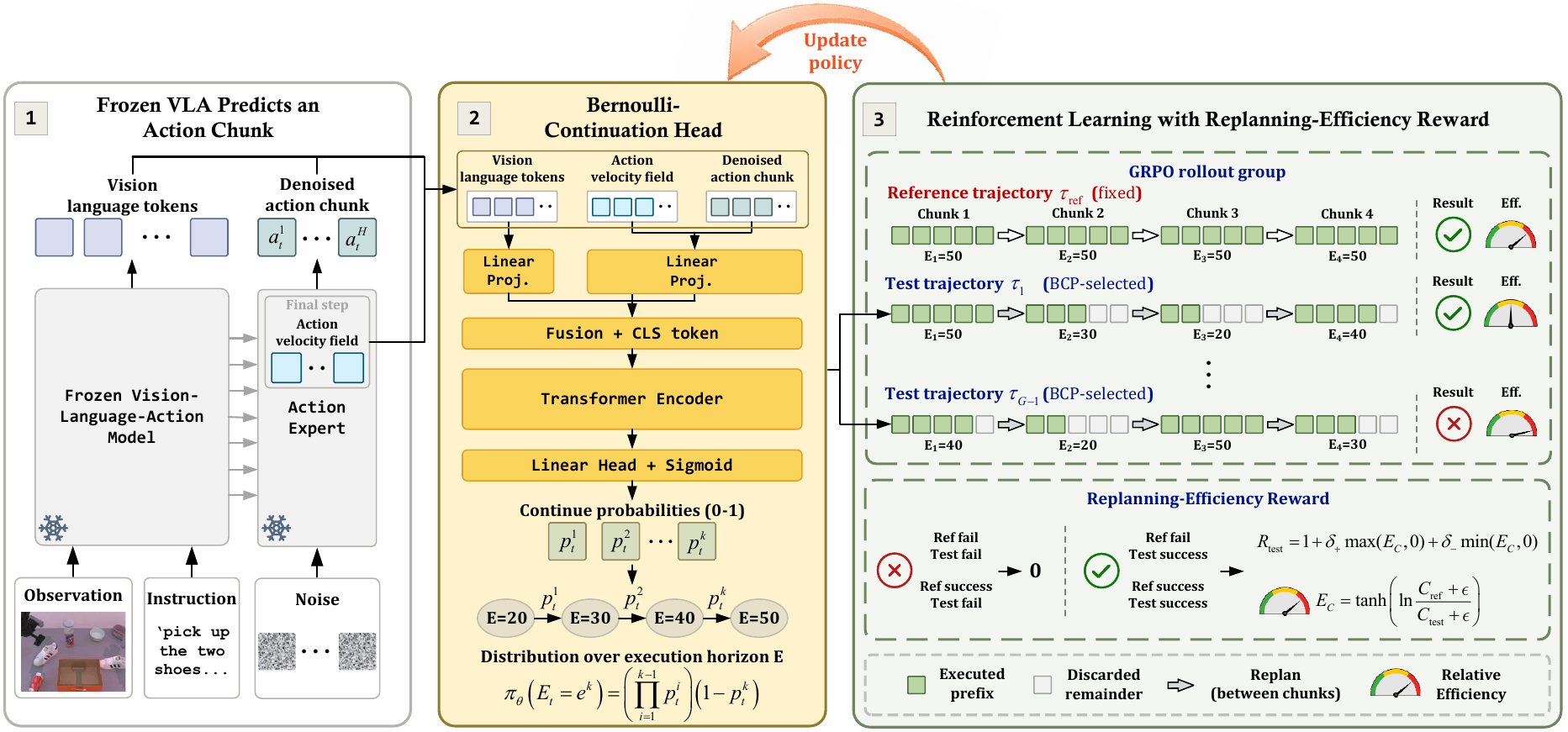}
\end{minipage}

\setlength{\abovecaptionskip}{4pt}
\setlength{\belowcaptionskip}{-10pt}
\caption{\textbf{Overview of Bernoulli-Continuation Policy learning}. Given a fixed-length action chunk predicted by a frozen VLA, BCP estimates continue probabilities to select an adaptive execution horizon, executes the corresponding action prefix, and is optimized with GRPO using the Replanning-Efficiency Reward.}
\label{fig:overview}
\end{figure*}

\section{Related Works}

\subsection{Adaptive Horizon Execution}
Action chunking is common in recent VLAs~\cite{zhao2023learning, kim2024openvla,chi2025diffusion,black2024pi_0,kim2025fine,bjorck2025gr00t,wang2026qwen}, which predict a fixed-length action sequence and execute it under a hand-fixed horizon. Recent works make this horizon adaptive, differing in the replanning signal and its cost. 
One line uses hand-designed test-time reliability proxies based on entropy or attention~\cite{liang2026adaptive,wang2026vla,zhu2026textsc}, which may not be directly optimized and often require sampling several chunks at each step.
A second line trains dedicated mechanisms, via cross-horizon consensus~\cite{jing2025mixture} or closed-loop verification~\cite{wang2026open,wang2026trust,pan2026vla}, which are effective but involve complex procedures. 
In contrast, we directly learn an execution-horizon distribution with a lightweight, plug-and-play policy. It predicts the execution horizon in a single forward pass and avoids complex procedures.

\subsection{RL for Robot Policy Optimization}
RL has become an important paradigm for improving VLA execution beyond imitation learning. SimpleVLA-RL~\cite{li2025simplevla}, $\pi_{\mathrm{RL}}$~\cite{chen2025pirl}, $\pi^*_{0.6}$~\cite{intelligence2025pi}, and VLA-RL~\cite{liu2026can} fine-tune the full VLA or its action generation.
Meanwhile, Q-chunking~\cite{li2026reinforcement} and AC3~\cite{yang2026actor} apply RL to improve exploration and sample efficiency while supporting high-dimensional continuous action chunks.
These methods improve \emph{how} actions are generated, but require updating a large action model, incurring substantial training cost.
We instead apply RL to an orthogonal target: a lightweight policy that decides \emph{when} to stop executing a chunk and replan, while keeping the base VLA frozen. 

\section{Method}
Figure~\ref{fig:overview} overviews our framework. We first formulate execution-horizon selection on a frozen chunk-based VLA (Sec.~3.1), then introduce the Bernoulli-Continuation Policy that models it as sequential continue-or-replan decisions (Sec.~3.2), and train it with reinforcement learning under a Replanning-Efficiency Reward, since per-chunk labels are unavailable (Sec.~3.3).

\subsection{Problem Formulation}
We consider a pretrained VLA model for robotic manipulation. Given the current observation $o_t$ and language instruction $\ell$, it predicts a fixed-length action chunk $\mathbf{a}_t = (a_t^1, a_t^2, \ldots, a_t^H)$, where $H$ denotes the prediction horizon. In standard horizon execution, the robot runs a fixed number of actions, the execution horizon $E$, before querying the VLA again. To make this adaptive, we define a discrete set of candidate execution horizons

\begin{equation}
\mathcal{E}=\left \{e^1,e^2,\ldots,e^M\right \}, \quad
1 \leq e^1 < \cdots < e^M \leq H,
\end{equation}

At each replanning step, an execution-horizon policy selects $E_t \sim \pi_\theta(\cdot \mid s_t)$ from $\mathcal{E}$, where $s_t$ is the information available for selection. The robot then executes the prefix $\mathbf{a}_t^{1:E_t}$ and queries the VLA again. Thus the VLA remains responsible for action generation, while the policy only decides how long each chunk is trusted before replanning.

\subsection{Bernoulli-Continuation Head}

Bernoulli-Continuation Head is a lightweight module attached to the frozen VLA. It reuses the representations already produced during action-chunk prediction, so decides the execution horizon without extra VLA forward pass. To judge whether the current chunk can still be trusted or a fresh observation is needed, it takes three VLA-derived inputs: visual-language tokens $\mathbf{F}_t \in \mathbb{R}^{N \times d_v}$ for scene and instruction context. For each action step $j$, the denoised action $\mathbf{a}_t^j \in \mathbb{R}^{d_a}$ and its final-step action-velocity feature $\mathbf{u}_t^j \in \mathbb{R}^{d_u}$, which expose motion-level cues about generated trajectory.

We concatenate the two action-level features and project them, together with the visual-language tokens, into a shared $d$-dimensional space:
\begin{equation}
\mathbf{h}_t^j = \mathrm{Proj}_a([\mathbf{u}_t^j;\mathbf{a}_t^j]),
\qquad
\hat{\mathbf{F}}_t = \mathrm{Proj}_v(\mathbf{F}_t).
\end{equation}
Prepending a learnable token $[\mathrm{CLS}]\in\mathbb{R}^{d}$, we form the input sequence $\mathbf{X}_t = [\,[\mathrm{CLS}],\ \hat{\mathbf{F}}_t,\ \mathbf{h}_t^1,\ldots,\mathbf{h}_t^H\,]$, feed it to a lightweight Transformer encoder so that $[\mathrm{CLS}]$ attends over both context and every action position, and map its output to $M-1$ logits $\mathbf{z}_t=(z_t^1,\ldots,z_t^{M-1})$.

Instead of a softmax classifier, we model horizon selection as an ordered sequence of Bernoulli continuation decisions. Each logit gives a continue probability
\begin{equation}
p_t^i=\sigma(z_t^i), \quad i=1,\ldots,M-1,
\end{equation}
where $p_t^i$ is the probability of continuing from horizon $e^i$ to $e^{i+1}$. Selecting horizon $e^k$ thus requires continuing through the first $k-1$ decisions and then stopping, giving
\begin{equation}
\pi_\theta(E_t=e^k\mid s_t)=
\begin{cases}
\left(\prod_{i=1}^{k-1}p_t^i\right)(1-p_t^k), & 1\leq k<M,\\
\prod_{i=1}^{M-1}p_t^i, & k=M.
\end{cases}
\end{equation}

This factorization matches the nested structure of chunk execution: a 40-step and a 50-step execution share the same first 40 actions and differ only in whether execution continues past step 40. By sharing the continue decisions across this common prefix, BCP imposes an ordinal inductive bias absent from a flat categorical policy, making a long horizon reachable only when the chunk is repeatedly judged reliable and letting adjacent horizons receive similar probabilities.

\subsection{RL with Replanning-Efficiency Reward}

The execution-horizon policy cannot be trained by supervised learning, since the optimal horizon for each chunk is not observable and a horizon decision can only be judged by its downstream effect on the rollout. The decisions are also temporally coupled, as the horizon chosen for the current chunk determines the next observation and hence later chunks and decisions. We therefore optimize BCP with reinforcement learning from trajectory-level rewards while keeping the base VLA frozen.

We train BCP with Group Relative Policy Optimization (GRPO)~\cite{shao2024deepseekmath}. For each task instance $\xi \sim \mathcal{D}$, $\pi_{\theta_{\mathrm{old}}}$ samples $G-1$ adaptive trajectories $\{\tau_i\}_{i=1}^{G-1}$, and we additionally rollout one reference trajectory $\tau_{\mathrm{ref}}$ under the fixed-horizon strategy to anchor the advantage. Each $\tau_i$ has $T_i$ replanning steps with horizons $E_{i,t}\sim\pi_{\theta_{\mathrm{old}}}(\cdot\mid s_{i,t})$, and receives a trajectory-level reward (defined below) broadcast to all its decisions. Only the $G-1$ adaptive trajectories update BCP, as $\tau_{\mathrm{ref}}$ has no BCP decisions. The policy is optimized by the clipped objective
\begin{equation}
 \resizebox{.99\linewidth}{!}{$
\begin{aligned}
J_{\mathrm{BCP}}(\theta)=&
\mathbb{E}_{\xi\sim\mathcal{D},\left \{\tau_i\right \}_{i=1}^{G-1}\sim\pi_{\theta_{\mathrm{old}}}}
\Bigg[
\frac{1}{\sum_{i=1}^{G-1}T_i}
\sum_{i=1}^{G-1}\sum_{t=1}^{T_i}\\
&
\min\left(
r_{i,t}(\theta)\hat{A}_i,\
\mathrm{clip}\left(r_{i,t}(\theta),1-\epsilon_{\mathrm{low}},1+\epsilon_{\mathrm{high}}\right)\hat{A}_i
\right)
\Bigg]
\end{aligned}
        $},
\end{equation}
where
\begin{equation}
r_{i,t}(\theta)=\frac{\pi_{\theta}\left(E_{i,t}\mid s_{i,t}\right)}{
\pi_{\theta_{\mathrm{old}}}\left(E_{i,t}\mid s_{i,t}\right)},
\hat{A}_i = \frac{ R_i-\mathrm{mean}\left( \mathcal{R} \right) }{\mathrm{std} \left( \mathcal{R} \right) } 
\end{equation}

where $\mathcal{R}=\left \{R_1,\ldots,R_{G-1},R_{\mathrm{ref}}\right \}$ collects the rewards of the $G-1$ adaptive trajectories and the reference. Here $r_{i,t}(\theta)$ is the importance ratio and $\hat{A}_i$ the group-normalized advantage. 
Including $R_{\mathrm{ref}}$ anchors the advantage to the fixed-horizon strategy, providing a stable reference for adaptive execution decisions.

A sparse binary success reward is insufficient here: optimizing success alone cannot distinguish a trajectory that succeeds with necessary replanning from one that succeeds by querying the VLA excessively, biasing the policy toward overly short horizons. The reward should thus encourage success while discouraging unnecessary VLA calls. To this end, our Replanning-Efficiency Reward instantiates $R_i$ by comparing each adaptive trajectory against the reference $\tau_{\mathrm{ref}}$, which anchors in-group efficiency. Let $S_i\in\{0,1\}$ denote whether $\tau_i$ succeeds, and $C_i, C_{\mathrm{ref}}$ the VLA calls of $\tau_i$ and $\tau_{\mathrm{ref}}$. We define the relative efficiency term as
\begin{equation}
\eta_i =
\tanh\left(
\log \frac{C_{\mathrm{ref}}}{C_i}
\right),
\end{equation}
where $\eta_i>0$ means adaptive execution uses fewer VLA calls than the fixed strategy, and $\tanh$ bounds its scale.

The final reward for an adaptive trajectory is
\begin{equation}
R_i =
S_i\left(
1+\delta_{+}\max(\eta_i,0)+\delta_{-}\min(\eta_i,0)
\right),
\end{equation}
where $\delta_{+}>\delta_{-}>0$ control the asymmetry. This keeps task success dominant, as a failed trajectory receives zero reward regardless of efficiency, so the policy cannot exploit short but unsuccessful executions. Among successful trajectories, it favors those using fewer VLA calls while only mildly penalizing extra replanning, encouraging BCP to replan before a critical stage when a fresh observation is needed, yet tolerating more replanning only when it preserves success.

The reward also improves group-based optimization. With binary success alone, a group is uninformative when all trajectories share an outcome. The efficiency term breaks such ties by ranking successful trajectories by VLA usage, so all-success groups still yield non-zero advantages, while the reference trajectory anchors the group so that failed adaptive trajectories still receive negative signals whenever the fixed strategy succeeds. Only groups in which the reference and all adaptive trajectories fail are discarded, which reduces the gradient sparsity caused by sparse task outcomes.

\begin{table*}[t]
\centering
\setlength{\abovecaptionskip}{3pt}
\vspace{-5pt}
\small
\setlength{\tabcolsep}{3pt}
\resizebox{0.98\textwidth}{!}{%
\begin{tabular}{l cc cc cc cc cc cc cc cc cc cc}
\toprule
\multirow{2}{*}{\textbf{Simulation Task}} &
\multicolumn{2}{c}{{ACT}} &
\multicolumn{2}{c}{{\textbf{ACT}$^{\dagger}$}}  &
\multicolumn{2}{c}{{$\pi_0$}} & \multicolumn{2}{c}{X-VLA} & \multicolumn{2}{c}{$\pi_{0.5}$} & \multicolumn{2}{c}{ABot-M0} & \multicolumn{2}{c}{\textbf{ABot-M0}$^{\dagger}$} & \multicolumn{2}{c}{Qwen-VLA} & \multicolumn{2}{c}{LingBot-VLA} & \multicolumn{2}{c}{\textbf{LingBot-VLA$^{\dagger}$}}\\
\cmidrule(lr){2-3} \cmidrule(lr){4-5} \cmidrule(lr){6-7} \cmidrule(lr){8-9} \cmidrule(lr){10-11} \cmidrule(lr){12-13} \cmidrule(lr){14-15} \cmidrule(lr){16-17} \cmidrule(lr){18-19} \cmidrule(lr){20-21}
 & Clean & Rand. & Clean & Rand. & Clean & Rand. & Clean & Rand. & Clean & Rand. & Clean & Rand. & Clean & Rand. & Clean & Rand. & Clean & Rand. & Clean & Rand. \\
\midrule
\textit{Beat Block Hammer}   & 58\% & 5\% & 73\% & 5\% & 79\% & 84\%  & 92\% & 88\%  & \textbf{96\%} & 93\% & 83\% & 79\% & 87\% & 86\% & \textbf{96\%} & \textbf{96\%}  & 82\% & 78\% & 91\% & 92\%\\
\textit{Blocks Ranking RGB}   & 1\% & 0\% & 3\% & 0\% & 80\% & 63\%  & 83\% & 83\% & 92\% & 85\% & 93\% & 89\% & 99\% & \textbf{98\%} & \textbf{100\%} & 95\% & 88\% & 88\% & 98\% & \textbf{98\%} \\
\textit{Blocks Ranking Size}   & 0\% & 0\% & 1\% & 0\%& 14\% & 5\%  & 67\% & 74\% & 49\% & 26\% & 61\% & 62\% & 74\% & 82\% & 70\% & 77\% & 70\% & 71\% & \textbf{88\% }& \textbf{89\%} \\
\textit{Click Alarmclock}   & 87\% & 8\% & 90\% & 8\% & 77\% & 68\%  & 99\% & 99\% & 98\% & 89\% & 75\% & 62\% & \textbf{100\%} & \textbf{100\%} & 85\% & 94\% & 83\% & 81\% & 91\% & 82\%\\
\textit{Hanging Mug}   & 13\% & 0\% & 21\% & 0\% & 14\% & 11\%  & 23\% & 27\% & 18\% & 17\% & 18\% & 24\% & 33\% & 34\% & 41\% & 44\% & 41\% & 36\% & \textbf{87\%} & \textbf{62\%} \\
\textit{Move Stapler Pad}   & 0\% & 0\% & 0\% & 0\% & 41\% & 24\%  & 78\% & 73\% & 56\% & 42\% & 48\% & 58\% & 71\% & 77\% & 64\% & 68\% & 75\% & 70\% & \textbf{82\%} & \textbf{78\%} \\
\textit{Place A2B Left}   & 5\% & 0\% & 10\% & 1\% & 43\% & 47\%  & 48\% & 49\% & 87\% & 82\% & 87\% & 88\% & \textbf{88\%} & \textbf{94\%} & 81\% & 89\% & 69\% & 67\% & 76\% & 81\%\\
\textit{Place A2B Right}   & 3\% & 0\% & 6\% & 0\% & 39\% & 34\%  & 36\% & 36\% & 87\% & 84\% & 81\% & 82\% & \textbf{88\%} & \textbf{93\%} & 87\% & 87\% & 81\% & 73\% & 79\% & 84\%\\
\textit{Place Can Basket}   & 14\% & 0\% & 28\% & 0\% & 55\% & 46\%  & 49\% & 52\% & 62\% & 62\% & 78\% & 66\% & 83\% & 74\% & 54\% & 74\% & 81\% & \textbf{83\%} & \textbf{84\%} & 81\%\\
\textit{Place Dual Shoes}   & 3\% & 0\% & 8\% & 0\% & 59\% & 51\%  & 79\% & \textbf{88\%} & 75\% & 75\% & 79\% & 76\% & \textbf{92\%} & 83\% & 85\% & 83\% & 81\% & 85\% & 90\% & \textbf{88\%} \\
\textit{Place Mouse Pad}   & 2\% & 0\% & 4\% & 0\% & 20\% & 20\%  & 70\% & 70\% & 60\% & 39\% & 64\% & 65\% & 79\% & 73\% & 73\% & 74\% & 86\% & 81\% & \textbf{92\%} & \textbf{88\%} \\
\textit{Place Object Scale}   & 0\% & 0\% & 2\% & 0\% & 57\% & 52\%  & 52\% & 74\% & 86\% & 80\% & 89\% & 86\% & \textbf{93\%} & \textbf{92\%} & 81\% & 85\%  & 80\% & 83\% & 91\% & 89\% \\
\textit{Turn Switch}   & 23\% & 6\% & 33\% & 12\% & 41\% & 42\%  & 40\% & 61\% & 62\% & 54\% & 63\% & 73\% & 71\% & \textbf{77\%} & 64\% & 72\%  & 60\% & 60\% & \textbf{72\%} & 73\% \\
\midrule
\textbf{Avg.  (13 tasks above) (\%)}  & \makecell[t]{16.08} & \makecell[t]{1.46} &   \makecell[t]{21.46 \\ \textbf{\textcolor{improvegreen}{(+5.38)}}} & \makecell[t]{2.00 \\ \textbf{\textcolor{improvegreen}{(+0.54)}}} & \makecell[t]{47.61} & \makecell[t]{42.08}  & \makecell[t]{62.77} & \makecell[t]{67.23}  & \makecell[t]{71.38} & \makecell[t]{63.69} & \makecell[t]{70.69} & \makecell[t]{70.00} &  \makecell[t]{81.38 \\ \textbf{\textcolor{improvegreen}{(+10.69)}} } & \makecell[t]{81.77 \\ \textbf{\textcolor{improvegreen}{(+11.77)}}} & \makecell[t]{75.46} & \makecell[t]{79.85} & \makecell[t]{75.15} & \makecell[t]{73.54} &   \makecell[t]{\textbf{86.23} \\ \textbf{\textcolor{improvegreen}{(+11.08)}}} &   \makecell[t]{\textbf{83.46} \\ \textbf{\textcolor{improvegreen}{(+9.92)}}}\\
\textbf{Avg.  (50 tasks) (\%)}  & \makecell[t]{34.50} & \makecell[t]{2.40} & \makecell[t]{-} & \makecell[t]{-} & \makecell[t]{65.92} & \makecell[t]{58.40}  & \makecell[t]{72.80} & \makecell[t]{72.84}  & \makecell[t]{82.74} & \makecell[t]{76.76} & \makecell[t]{85.96} & \makecell[t]{85.40} & \makecell[t]{-} & \makecell[t]{-} & \makecell[t]{86.10} & \makecell[t]{87.20} & \makecell[t]{89.88} & \makecell[t]{88.78} &   \makecell[t]{\textbf{93.94} \\ \textbf{\textcolor{improvegreen}{(+4.06)}}} &   \makecell[t]{\textbf{92.84} \\ \textbf{\textcolor{improvegreen}{(+4.06)}}}\\
\bottomrule
\end{tabular}%
}
\caption{Success rate comparison on RoboTwin 2.0. The table lists 13 low-success tasks where the LingBot-VLA obtains below 90\% success rate under the Clean setting, together with average results over all 50 tasks. ${\dagger}$ denotes the addition of BCP.}
\label{tab:robotwin_full}
\vspace{-5pt}
\end{table*}

\section{Experiments}

\subsection{Experimental Settings}
\paragraph{Simulation Settings.} We conduct experiments on RoboTwin 2.0~\cite{chen2025robotwin}, LIBERO~\cite{liu2023libero} and LIBERO-PRO~\cite{zhou2025liberopro} to evaluate the effectiveness and generalization ability of the proposed method. On RoboTwin 2.0, BCP is trained under the Clean setting and evaluated under both Clean and Randomized settings. 
We use LingBot-VLA w/o depth~\cite{wu2026pragmatic}, ABot-M0~\cite{yang2026abot}, and ACT~\cite{zhao2023learning} as base policies. LingBot-VLA and ABot-M0 represent large-scale VLA models, whereas ACT serves as a lightweight, earlier-generation chunk-based policy. All three policies use a default prediction horizon and execution horizon of 50 steps.
BCP only learns an additional continuation head on top of the frozen VLA. The candidate execution horizons are set to ({15, 20, 25,..., 50}), and the policy is implemented with a 2-layer Transformer encoder. For the Reference-Relative Reward, we set $\delta_{+}=0.7$ and $\delta_{-}=0.3$. We train BCP with GRPO on 8 NVIDIA A100 GPUs (40GB) for 300 steps, using a group size of $G=8$. During rollout stage, we use 256 parallel environments and set the number of rollout epochs to 4. During training stage, we use a global batch size of 512 and perform 2 update epochs.
All simulation experiments are implemented on top of RLinf~\cite{yu2026rlinf}, an open-source framework for reinforcement learning of VLA.
\paragraph{Real-World Settings.} We conduct two manipulation tasks on the AGIBOT G1 using LingBot-VLA as the base policy. For each task, we collect 250 task-specific teleoperated trajectories and jointly train the SFT policy with 2,000 pre-collected general-purpose grasping trajectories for 40,000 steps. We then collect 128 real-world trajectories for reinforcement learning to train BCP while keeping the base policy frozen. For evaluation, all compared methods are tested on the same set of 50 seeds for each task with three trials.

\subsection{Benchmark Results}
\paragraph{Results on RoboTwin 2.0.} Table \ref{tab:robotwin_full} reports the main results on RoboTwin 2.0. To better examine whether adaptive execution can improve difficult manipulation scenarios, we first focus on 13 low-success tasks where the original LingBot-VLA achieves a success rate below 90\% under the Clean setting. With the proposed BCP, LingBot-VLA improves from 75.15\% to 86.23\% on these 13 tasks, yielding an absolute gain of \textbf{11.08\%}. This suggests that many failures of the fixed execution strategy arise not from poor action-chunk prediction, but from the misalignment between fixed replanning boundaries and critical manipulation stages.
Across all 50 RoboTwin 2.0 tasks, LingBot-VLA + BCP further improves the average success rate from 89.88\% to \textbf{93.94\%}, achieving a \textbf{4.06\% gain} and the state-of-the-art performance among all VLA methods. Similar trends are also observed with ABot-M0 and ACT.
We further evaluate whether the learned execution-horizon policy can generalize beyond the training distribution. Although BCP is trained only under the Clean setting, we directly test it under the \textbf{Randomized} setting. On the 13 low-success tasks, LingBot-VLA + BCP improves the success rate from 73.54\% to 83.46\%, yielding a \textbf{9.92\%} absolute gain. Across all 50 RoboTwin 2.0 tasks, BCP improves the average success rate from 88.78\% to \textbf{92.84\%}, with a \textbf{4.06\%} gain, and again achieves the state-of-the-art performance.
These results indicate that BCP does not merely overfit to the specific visual conditions seen during training. Instead, the improvement suggests that the policy captures task-stage-level knowledge about when a predicted action chunk should be trusted and when replanning is necessary.

\begin{figure*}[t]
\centering

\begin{subfigure}[c]{0.98\linewidth}
    \centering
    \includegraphics[width=\linewidth]{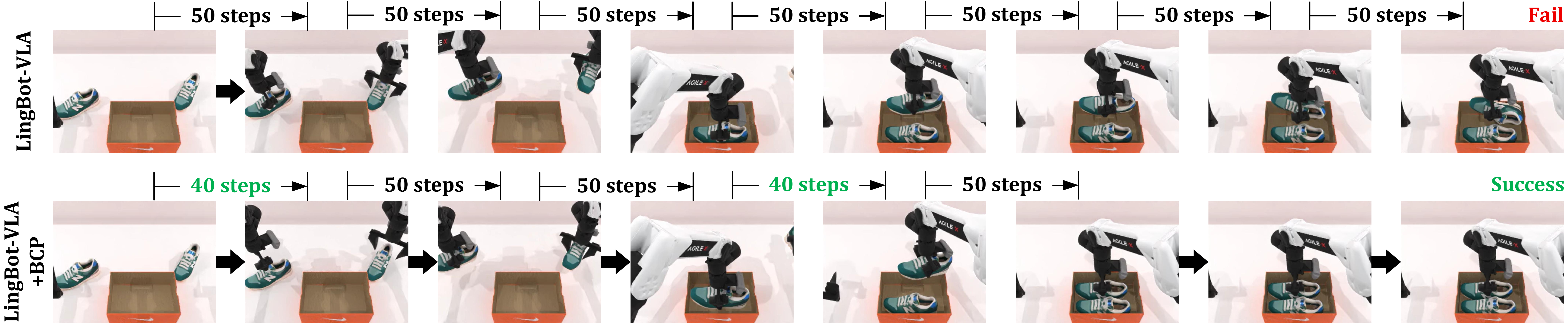}
    \caption{\emph{Place Dual Shoes} (Clean settings)}
    \label{fig:3a}
\end{subfigure}

\vspace{4pt}

\begin{subfigure}[c]{0.98\linewidth}
    \centering
    \includegraphics[width=\linewidth]{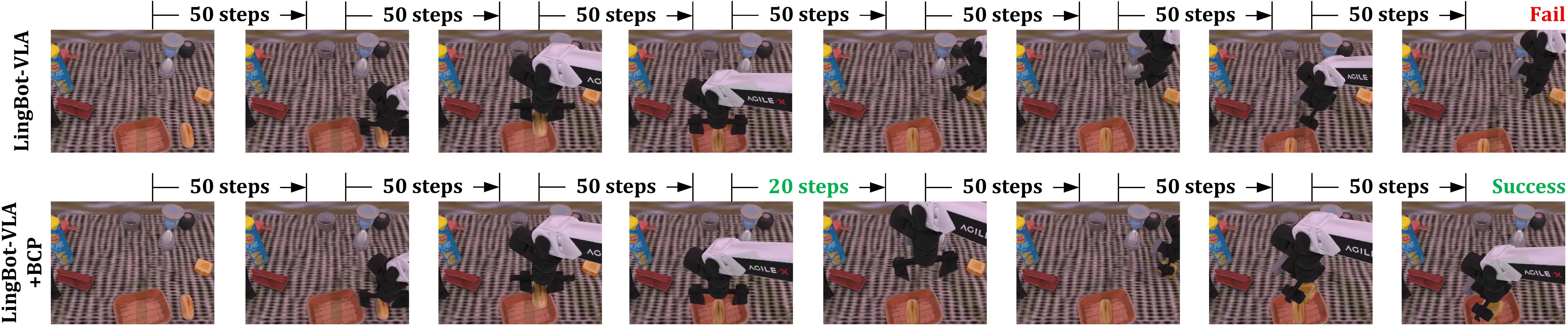}
    \caption{\emph{Place Bread Basket} (Randomized setting)}
    \label{fig:3b}
\end{subfigure}

\setlength{\abovecaptionskip}{4pt}
\setlength{\belowcaptionskip}{-10pt}
\caption{Qualitative examples on RoboTwin 2.0. BCP adaptively replans before critical manipulation stages, refreshing the observation instead of acting on a stale chunk, and improves task execution over the fixed strategy.}
\label{fig:vis}
\end{figure*}

\vspace{-5pt}

\begin{table}[t]
    \centering
    \setlength{\abovecaptionskip}{3pt}
    \setlength{\belowcaptionskip}{-10pt}
    \footnotesize
    \setlength{\tabcolsep}{3pt}
    \begin{tabular}{lccccc}
        \toprule
        \multicolumn{6}{c}{LIBERO} \\
        \midrule
        Method & Spatial & Object & Goal & Long & Avg. \\
        \midrule
        $\pi_{0.5}$   & 98.5\% & 98.7\% & 98.1\% & 92.5\% & 97.0\% \\
        $\pi_{0.5}$ + AAC & 99.1\%  & 99.2\%  & 98.0\%  & 95.2\%  & 97.9\%  \\
        $\pi_{0.5}$ + AutoHorizon & 99.1\%  & 99.2\%  & 97.5\%  & 91.6\%  & 96.9\%  \\
        $\pi_{0.5}$ + BCP & \textbf{99.6\%} & \textbf{99.8\%}
        & \textbf{99.2\%} & \textbf{96.2\%} & \textbf{98.7\%} \\
        \midrule
        \multicolumn{6}{c}{LIBERO-PRO} \\
        \midrule
        Method & $\times 0.2$ & $\times 0.3$ & $\times 0.4$ & \multicolumn{2}{c}{Avg.}\\
        \midrule
        $\pi_{0.5}$ & 53.2\%  & 29.9\%  & 9.5\%  & \multicolumn{2}{c}{30.9\%} \\
        $\pi_{0.5}$ + AAC & \textbf{57.4\%} & 35.3\%  & 11.8\%  & \multicolumn{2}{c}{34.8\%} \\
        ${\pi_{0.5}}$ + BCP & 57.0\%  &\textbf{39.4\%} &\textbf{16.8\%}
        & \multicolumn{2}{c}{\textbf{37.7\%}} \\
        \bottomrule
    \end{tabular}
    \caption{
        Success rates (\%) of different execution strategies with
        $\pi_{0.5}$ on LIBERO and LIBERO-PRO.
    }
    \label{tab:libero_results}
\end{table}

\paragraph{Results on LIBERO and LIBERO-PRO.} We further evaluate BCP with $\pi_{0.5}$ on LIBERO and LIBERO-PRO. As shown in Table \ref{tab:libero_results}, BCP achieves the best performance across all four LIBERO task suites, improving the average success rate of $\pi_{0.5}$ from 97.0\% to 98.7\% and outperforming both AAC~\cite{liang2026adaptive} and AutoHorizon~\cite{wang2026vla}. On LIBERO-PRO, following the evaluation protocol of AAC, we evaluate position perturbations of different magnitudes on the Object suite. BCP improves the average success rate from 30.9\% to 37.7\%, corresponding to a 2.9\% gain over AAC. The consistent improvements, particularly under stronger position perturbations, demonstrate that BCP learns a robust execution-horizon policy that generalizes well to challenging environmental variations.

\subsection{Runtime Efficiency}

To verify that the performance gains of BCP do not come at the cost of substantially higher runtime overhead, we compare LingBot-VLA and LingBot-VLA + BCP on all 50 RoboTwin 2.0 tasks under the Clean setting. We first measure the average inference time of a single VLA query on an NVIDIA A100 GPU. Since the continuation head is lightweight, it only increases the per-query inference time from 938.10 ms to 940.13 ms, introducing an additional overhead of 2.03 ms.
We further record the average number of VLA calls and the average number of executed control steps per episode. For the ALOHA-AgileX embodiment, whose control frequency is 50 Hz, the overall runtime is estimated by combining the model inference cost and the robot execution time.
As shown in Table \ref{tab:time_cost}, BCP slightly increases the average number of VLA calls from 5.382 to 5.614 due to adaptive replanning. However, it reduces the average number of executed steps from 269.075 to 248.102. As a result, the estimated runtime decreases from 10.43 s to 10.24 s. These results show that BCP introduces negligible per-query inference overhead and preserves the runtime efficiency of action chunking, while enabling selective replanning to improve task execution.

\begin{table}[h]
\setlength{\abovecaptionskip}{4pt}
\setlength{\belowcaptionskip}{-15pt}
\centering
\footnotesize
\setlength{\tabcolsep}{3pt}
\begin{tabular}{lcccc}
    \toprule
    Method & \makecell{VLA Infer\\Time (ms)} & \makecell{VLA \\ Calls} & \makecell{Exec. \\ Steps} & Runtime (s)\\
    \midrule
    LingBot-VLA      & 938.10 & 5.382 & 269.075 & 10.43 \\
    \textbf{LingBot-VLA +BCP} & 940.13 & 5.614 & 248.102 & 10.24 \\
    \bottomrule
\end{tabular}
\caption{Runtime cost comparison on 50 RoboTwin 2.0 tasks under the Clean setting. Although BCP issues slightly more VLA calls, it executes fewer control steps, so the overall runtime is lower.}
\label{tab:time_cost}
\end{table}

\subsection{Comparison with Execution-Horizon Strategies}

\begin{figure}[h]
\centering
\begin{minipage}[c]{0.9\linewidth}
\centering
\includegraphics[width=1.0\textwidth]{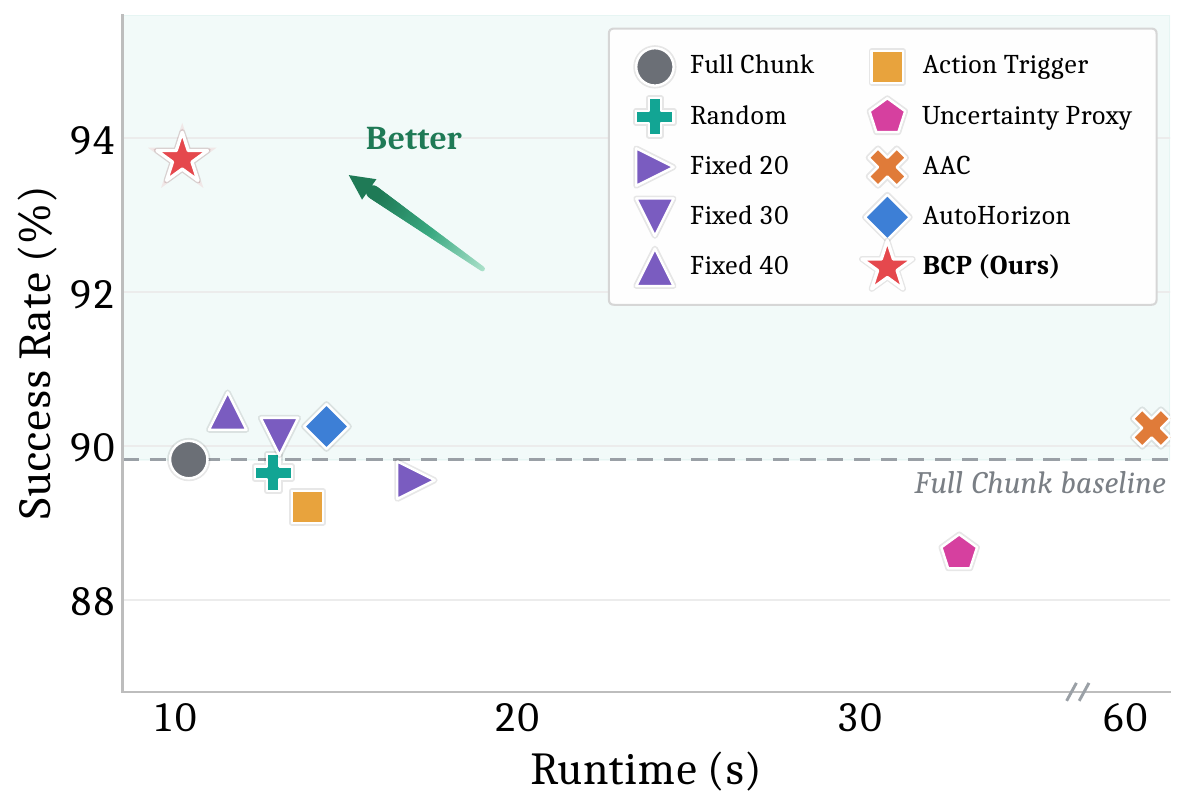}
\end{minipage}

\setlength{\abovecaptionskip}{4pt}
\setlength{\belowcaptionskip}{-20pt}
\caption{SR--runtime trade-off of execution-horizon strategies on 50 RoboTwin 2.0 tasks under the Clean setting.}
\label{fig:compare}
\end{figure}

\begin{figure*}[h]
\centering
\captionsetup[subfigure]{skip=1pt}

\begin{subfigure}[c]{0.98\linewidth}
    \centering
    \includegraphics[width=\linewidth]{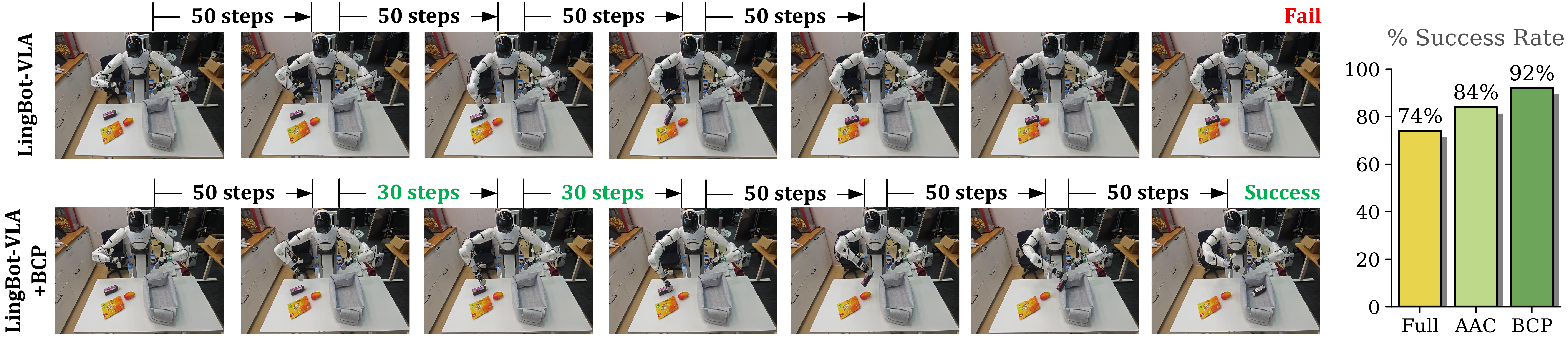}
    \caption{\emph{Grasping Bottle}: Pick up the grape juice bottle on the tabletop with the right arm and place it into the felt bag.}
    \label{fig:5a}
\end{subfigure}

\vspace{4pt}

\begin{subfigure}[c]{0.98\linewidth}
    \centering
    \includegraphics[width=\linewidth]{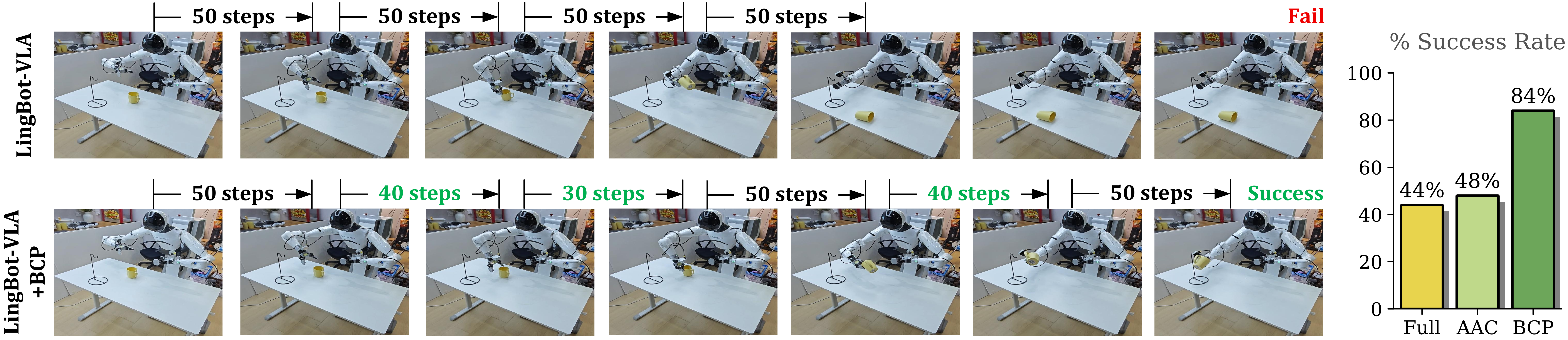}
    \caption{\emph{Hanging Mug}: Use the right arm to pick the mug and hang it onto the rack.}
    \label{fig:5b}
\end{subfigure}

\setlength{\abovecaptionskip}{4pt}
\setlength{\belowcaptionskip}{-15pt}
\caption{Real-world evaluation on the AGIBOT G1.}
\label{fig:real-robot}
\end{figure*}

We compare BCP with 8 execution-horizon strategies built upon LingBot-VLA, including Random, Fixed 20, Fixed 30, Fixed 40, Action Trigger~\cite{wang2026vla}, Uncertainty Proxy~\cite{wang2026vla}, AAC~\cite{liang2026adaptive}, and AutoHorizon~\cite{wang2026vla}, as shown in Figure \ref{fig:compare}. The results first show that a shorter execution horizon is not inherently better: Fixed 20 fails to outperform the full-chunk baseline because it still imposes a uniform replanning schedule that can remain misaligned with critical manipulation stages. Moreover, Uncertainty Proxy and AAC incur substantially higher runtime because they require sampling multiple action chunks to estimate uncertainty or entropy. Overall, nearly all competing strategies provide only marginal gains or even degradation, highlighting the difficulty of selecting suitable replanning boundaries. In contrast, BCP improves the average success rate by 4.06\%. Benefiting from the Replanning-Efficiency Reward, which penalizes inefficient horizon 
decisions, BCP replans only when necessary and consequently achieves the lowest runtime among all evaluated methods.

\subsection{Ablation studies}
Table \ref{tab:bcp_ablation} analyzes the contribution of each component. The SFT policy achieves a success rate of only 41\%, highlighting the substantial headroom for adaptive horizon execution. Directly fine-tuning the action expert with RL (Fixed-horizon) increases the success rate to 79\%, but requires significantly higher computational overhead. Training only a lightweight head with a conventional softmax classifier (Softmax), achieves a comparable success rate of 78\% with a much smaller trainable module. Replacing the softmax classifier with Bernoulli-Continuation (BC) Head further improves the success rate to 83\%, indicating that explicitly modeling the ordinal, prefix-sharing structure helps. Finally, incorporating the Replanning-Efficiency Reward (RER) raises the success rate to 87\%, achieving the best performance. This demonstrates that jointly accounting for task success and replanning efficiency provides a more informative learning signal.

\begin{table}[t]
    \centering
    \setlength{\abovecaptionskip}{4pt}
    \setlength{\belowcaptionskip}{-15pt}
    \small
    \setlength{\tabcolsep}{2pt}
    \begin{tabular}{lcc}
        \toprule
        Method & Trainable Params. & Success Rate \\
        \midrule
        SFT      & -      & 41\% \\
        \midrule
        RL (Fixed-horizon) & 442.803 M   & 79\% \textbf{\textcolor{improvegreen}{(+38\%)}} \\
        \midrule
        RL (Softmax Head) & 16.394 M & 78\% \textbf{\textcolor{improvegreen}{(+37\%)}} \\
        RL (BC Head) & 16.393 M   & 83\% \textbf{\textcolor{improvegreen}{(+42\%)}} \\
        RL (BC Head) + RER  & 16.393 M     & 87\% \textbf{\textcolor{improvegreen}{(+46\%)}} \\
        \bottomrule
    \end{tabular}
    \caption{Ablation study on the \emph{Hanging Mug} task under the Clean setting of RoboTwin 2.0.}
    \label{tab:bcp_ablation}
\end{table}

\subsection{Visualization}

Figure \ref{fig:vis} provides qualitative examples to illustrate how BCP improves execution by selecting state-dependent replanning boundaries. In the \emph{Place Dual Shoes} task, when placing the second shoe, BCP selects a 40-step horizon, triggering an additional replanning step before the final placement. With the updated observation, the robot obtains a corrected placement angle and places the shoe properly inside the box, whereas the fixed 50-step strategy moves it toward the edge.
In the \emph{Place Bread Basket} task, the original LingBot-VLA fails when grasping the second bread because its fixed replanning schedule does not set a replanning boundary before the manipulation stage. In contrast, BCP divides the movement toward the second bread into two chunks, with the second chunk terminating immediately before grasping. This 
timely replanning allows the robot to generate a more accurate grasping motion.

\vspace{-3pt}

\subsection{Real-Robot Experiments}
Figure \ref{fig:real-robot} presents results on two real-world manipulation tasks. For the \emph{Grasping Bottle} task, the fixed full-chunk strategy achieves a success rate of 74\%. BCP further raises the success rate to 92\%. The main challenge is that the smooth bottle surface makes the grasp sensitive to small pose errors. By adaptively shortening the execution horizon during the approach and manipulation stages, BCP refreshes the observation and obtains a more accurate grasping pose, thereby preventing the bottle from slipping. For the \emph{Hanging Mug} task, the improvement is more pronounced: BCP achieves an 84\% success rate, compared with 44\% for full-chunk execution and 48\% for AAC. As illustrated in the execution sequence, BCP triggers replanning before the hanging stage and generates a more precise hanging motion.

\section{Conclusion}
In this paper, we showed that a fixed execution horizon turns replanning into a task-agnostic periodic schedule that can be misaligned with critical manipulation stages, and that this timing, not horizon length alone, governs success. To address it, we proposed the Bernoulli-Continuation Policy, which formulates horizon selection as a chain of continue-or-replan decisions and is trained with reinforcement learning under a Replanning-Efficiency Reward, keeping the base VLA frozon. Across RoboTwin 2.0, LIBERO and a real robot, BCP consistently improves the fixed-horizon base policy without adding inference cost, with the largest gains on timing-sensitive, low-success task.

\section*{Acknowledgments}
We thank Wenbo Li, Jiaxing Qiu, and Chao Yu from the RLinf development team. We also thank Bohan Li for his helpful discussions on reinforcement learning.

\bibliography{2027}

\end{document}


\maketitle

\section*{Contents}

\manualtocsection{setup}
{1\quad Experimental Settings}

\manualtocsubsection{real-setup}
{1.1\quad Real-World Experimental Settings}

\manualtocsubsection{sim-setup}
{1.2\quad Simulation Experimental Settings}

\manualtocsection{additional-results}
{2\quad Additional Experimental Results}

\manualtocsubsection{phase-distribution}
{2.1\quad Distribution of Best Phases in the Phase-Shift Experiment}

\manualtocsubsection{task-results}
{2.2\quad Detailed Task-Wise Results on RoboTwin 2.0}

\manualtocsubsection{runtime-analysis}
{2.3\quad Runtime Efficiency Analysis}

\manualtocsubsection{hyperparameter-study}
{2.4\quad Hyperparameter}

\manualtocsection{limitations}
{3\quad Limitations and Future Work}

\pdfanchor{setup}
\section{Experimental Settings}
\label{setup}

\pdfanchor{real-setup}
\subsection{Real-World Experimental Settings}
\label{real-setup}

Table~\ref{tab:hyperparam_real_sft} summarizes the configurations used for real-world supervised fine-tuning (SFT) and reinforcement learning (RL) training. 
All real-world experiments are conducted on the AGIBOT G1. For visual perception, we use three cameras: one top-view camera and two wrist-mounted cameras.
The robot has 14 arm joints and two end effectors. 
We adopt LingBot-VLA-4B as the base model and freeze its vision encoder during SFT. 
The SFT stage uses 250 task-specific teleoperated trajectories together with approximately 2,000 pre-collected general-purpose grasping trajectories. 

For adaptive action-chunk execution, the Bernoulli-Continuation Policy (BCP) selects from five candidate execution horizons, $\{30,35,40,45,50\}$.
Its continuation head consists of a 2-layer Transformer with 8 attention heads.
BCP is subsequently optimized using GRPO with a group size of 8.
For Replanning-Efficiency Reward (RER), we set $\delta_{+}=0.7$ and $\delta_{-}=0.3$. 
The RL stage uses 128 training trajectories and 4 update epochs.

Figure~\ref{fig:real_world_tasks} presents the physical setups and representative evaluation instances of the two real-world tasks. 
The tasks are designed to evaluate adaptive execution in manipulation stages that require accurate grasping or contact alignment.

\begin{table}[H]
\centering
\setlength{\abovecaptionskip}{3pt}
\setlength{\belowcaptionskip}{-10pt}
\resizebox{\linewidth}{!}{%
\begin{tabular}{lc}
\toprule
\textbf{Parameters} & \makecell{\textbf{LingBot-VLA} \\ (Real-world dataset)} \\
\midrule
\multicolumn{2}{l}{\textit{\textbf{Real Robot}}} \\
Type                          & AGIBOT G1 \\
Cameras                          & \makecell[l]{camera\_top, camera\_wrist\_left, \\ camera\_wrist\_right} \\
Arm joints / effector            & 14 / 2 \\
\midrule
\multicolumn{2}{l}{\textit{\textbf{Base Model}}} \\
Type                       & LingBot-VLA-4B \\
Freeze vision encoder            & True \\
Precision (mixed)          & FP32 \\
\midrule
\multicolumn{2}{l}{\textit{\textbf{SFT Training}}} \\
Training trajectories            & \makecell{250 (task-specific) \\ + 2,000 (general-purpose)} \\
Training steps               & 40000 \\
Global batch size                & 256 \\
Learning rate                    & 5e-5 \\
LR decay style                   & Constant \\
\midrule
\multicolumn{2}{l}{\textit{\textbf{Bernoulli-Continuation Head}}} \\
Num classes                      & 5  \\
Execute steps                    & \{30, 35, 40, 45, 50\}  \\
Transformer layers / heads       & 2 / 8  \\
Init mode                        & normal  \\
Loss coef                 & 1.0  \\
Sampling temperature             & 1.0  \\
\midrule
\multicolumn{2}{l}{\textit{\textbf{RL Training}}} \\
Advantage type                   & GRPO \\
Group size                       & 8  \\
Clip ratio (low / high)          & 0.2 / 0.28  \\
Entropy bonus                    & 0.05  \\
RER $\delta_+ / \delta_-$ & 0.7 / 0.3  \\
Training trajectories            & 128 \\
Update epochs                  & 4 \\
Global batch size                & 512 \\
Learning rate                    & 5e-5 \\
Weight decay                     & 0.001  \\
Grad clip                        & 1.0 \\
LR decay style                   & Constant \\
\bottomrule
\end{tabular}%
}
\caption{Hyperparameters for real-world SFT and RL training with LingBot-VLA.}
\label{tab:hyperparam_real_sft}
\end{table}

\begin{figure*}[t]
    \centering

    \begin{subfigure}[t]{0.99\linewidth}
        \centering

        \sbox{\firstrowref}{%
            \includegraphics[width=0.235\linewidth]
            {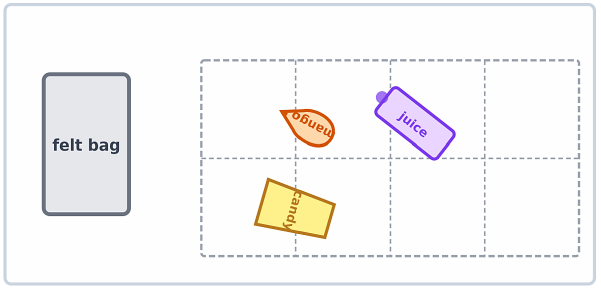}%
        }

        \begin{tabular}{@{}c@{}c@{}c@{}c@{}}

            \makebox[0.245\linewidth][c]{%
                \includegraphics[
                    width=\wd\firstrowref,
                    height=\dimexpr\ht\firstrowref+\dp\firstrowref\relax
                ]{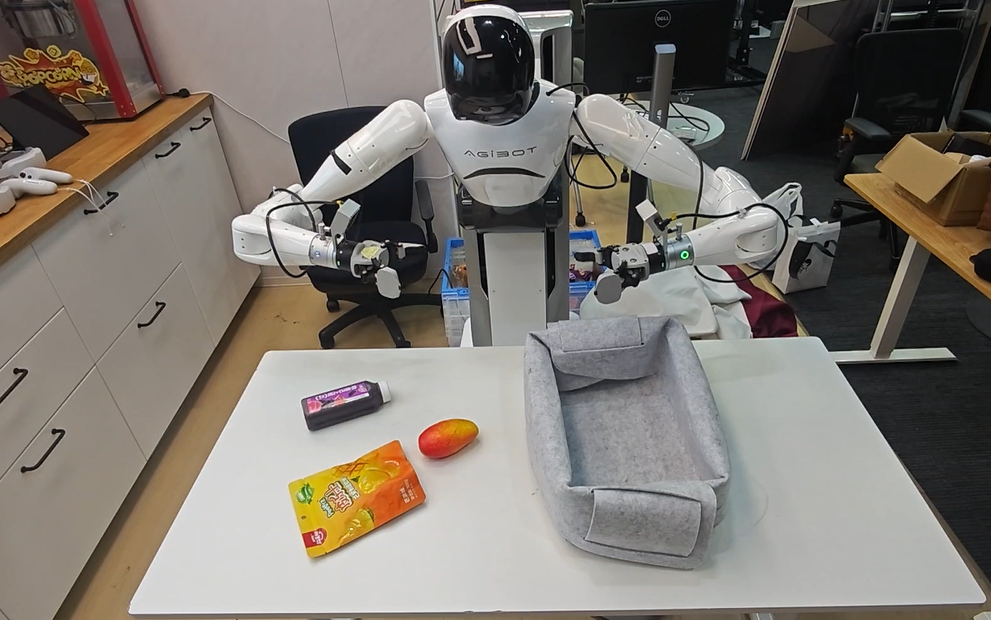}%
            }
            &

            \makebox[0.245\linewidth][c]{%
                \usebox{\firstrowref}%
            }
            &
            \makebox[0.245\linewidth][c]{%
                \includegraphics[width=0.235\linewidth]
                {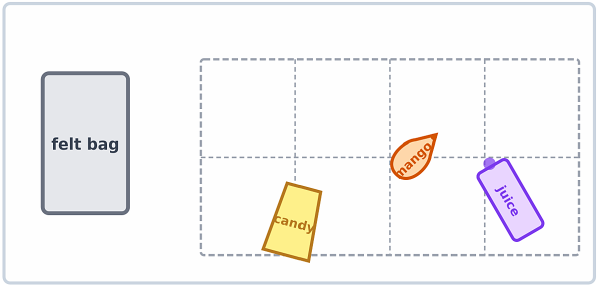}%
            }
            &
            \makebox[0.245\linewidth][c]{%
                \includegraphics[width=0.235\linewidth]
                {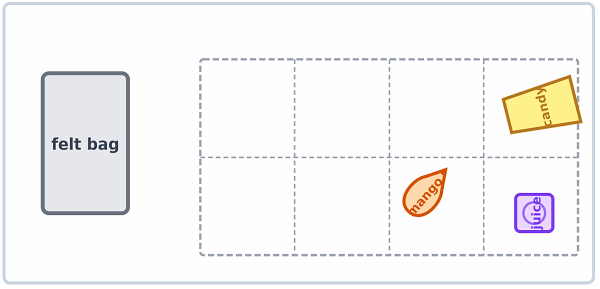}%
            }

        \end{tabular}

        \caption{\emph{Grasping Bottle}: Pick up the grape juice bottle on the tabletop with the right arm and place it into the felt bag.}
        \label{fig:first_row}
    \end{subfigure}

    \vspace{4pt}

    \begin{subfigure}[t]{0.99\linewidth}
        \centering

        \sbox{\secondrowref}{%
            \includegraphics[width=0.235\linewidth]
            {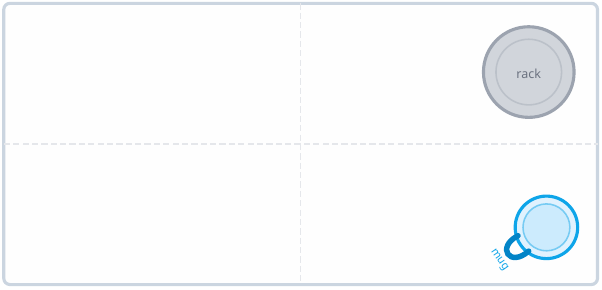}%
        }

        \begin{tabular}{@{}c@{}c@{}c@{}c@{}}

            \makebox[0.245\linewidth][c]{%
                \includegraphics[
                    width=\wd\secondrowref,
                    height=\dimexpr\ht\secondrowref+\dp\secondrowref\relax
                ]{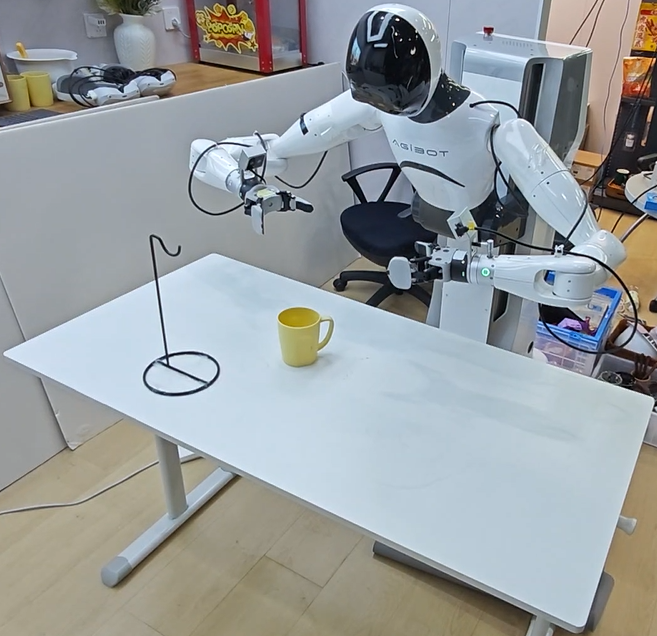}%
            }
            &
 
            \makebox[0.245\linewidth][c]{%
                \usebox{\secondrowref}%
            }
            &
            \makebox[0.245\linewidth][c]{%
                \includegraphics[width=0.235\linewidth]
                {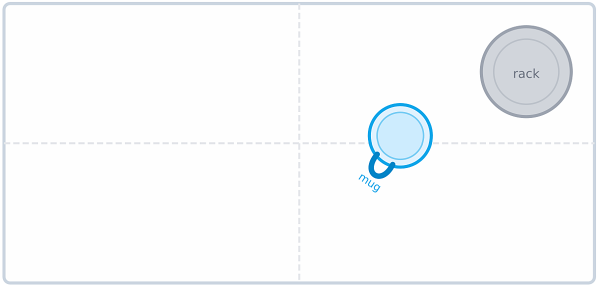}%
            }
            &
            \makebox[0.245\linewidth][c]{%
                \includegraphics[width=0.235\linewidth]
                {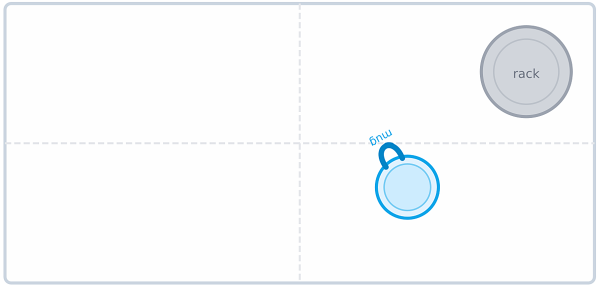}%
            }

        \end{tabular}

        \caption{\emph{Hanging Mug}: Use the right arm to pick the mug and hang it onto the rack.}
        \label{fig:second_row}
    \end{subfigure}

    \setlength{\abovecaptionskip}{4pt}
    \setlength{\belowcaptionskip}{-10pt}
    \caption{Real-world task settings and some evaluation instances for \emph{Grasping Bottle} and \emph{Hanging Mug} on the AGIBOT G1.}
    \label{fig:real_world_tasks}
\end{figure*}

\begin{itemize}
    \item \textbf{\emph{Grasping Bottle}.}
    The robot is required to pick up the grape juice bottle from the tabletop using its right arm and place it into the felt bag. 
    The primary challenge is that the bottle has a smooth surface. 
    Even a small error in the predicted grasp pose may result in an unstable grasp, causing the bottle to slip from the gripper during lifting or transportation.

    \item \textbf{\emph{Hanging Mug}.}
    The robot is required to pick up the mug using its right arm and hang it onto the rack. 
    Successful execution requires precise alignment between the mug handle and the rack hook. 
    In particular, the relative positional error must remain within 2\,cm. Otherwise, the handle may miss the hook or collide with the rack.
\end{itemize}

\begin{table*}[h]
\centering
\setlength{\abovecaptionskip}{6pt}
\setlength{\belowcaptionskip}{-15pt}
\resizebox{0.9\linewidth}{!}{%
\begin{tabular}{lcccc}
\toprule
\textbf{Parameters} & \makecell{\textbf{LingBot-VLA} \\ (RoboTwin 2.0)} & \makecell{\textbf{ABot-M0} \\ (RoboTwin 2.0)} & \makecell{\textbf{ACT} \\ (RoboTwin 2.0)} & \makecell{$\boldsymbol{\pi_{0.5}}$ \\ (LIBERO)} \\
\midrule
\multicolumn{5}{l}{\textit{\textbf{Base Model}}} \\
Prediction horizon                    & 50 & 50 & 50 & 10 \\
Denoise steps                    & 10 & 4 & - & 10 \\
Noise method                     & Euler ODE & Euler ODE & - & Euler ODE \\
Precision (mixed)                & BF16 & BF16 & FP32 & BF16 \\
\midrule
\multicolumn{5}{l}{\textit{\textbf{Bernoulli-Continuation Head}}} \\
Num classes                      & 8 & 8 & 8 & 6 \\
Execute steps                    & \{15, 20, 25,..., 50\} & \{15, 20, 25,..., 50\} & \{15, 20, 25,..., 50\} & \{2, 4, 5, 6, 8, 10\} \\
Transformer layers / heads       & 2 / 8 & 2 / 8 & 2 / 8 & 2 / 8 \\
Init mode                        & normal & normal & normal & normal \\
Loss coef                 & 1.0 & 1.0 & 1.0 & 1.0 \\
Sampling temperature             & 1.0 & 1.0 & 1.0 & 1.0 \\
\midrule
\multicolumn{5}{l}{\textit{\textbf{Rollout / Environment}}} \\
Rollout epochs                   & 4 & 16 & 4 & 8 \\
Train parallel environments      & 256 & 64 & 256 & 64 \\
Eval environments       & 100 & 100 & 100 & 500 \\
\midrule
\multicolumn{5}{l}{\textit{\textbf{Reinforcement Learning Algorithm}}} \\
Advantage type                   & GRPO & GRPO & GRPO & GRPO \\
Group size                       & 8 & 8 & 8 & 8 \\
Clip ratio (low / high)          & 0.2 / 0.28 & 0.2 / 0.28 & 0.2 / 0.28 & 0.2 / 0.28 \\
Entropy bonus                    & 0.05 & 0.05 & 0.05 & 0.05 \\
RER $\delta_+ / \delta_-$ & 0.7 / 0.3 & 0.7 / 0.3 & 0.7 / 0.3 & 0.7 / 0.3 \\
\midrule
\multicolumn{5}{l}{\textit{\textbf{Training}}} \\
Total training steps                     & 300 & 300 & 300 & 300 \\
Global batch size                & 512 & 512 & 512 & 512 \\
Micro batch size                 & 4 & 2 & 4 & 4 \\
Update epochs                    & 2 & 2 & 2 & 4 \\
learning rate                         & 1e-4 & 1e-4 & 1e-4 & 1e-4 \\
Weight decay                     & 0.001 & 0.001 & 0.001 & 0.001 \\
Grad clip                        & 1.0 & 1.0 & 1.0 & 1.0 \\
LR scheduler                     & Constant & Constant & Constant & Constant \\
\bottomrule
\end{tabular}%
}
\caption{Simulation hyperparameters across base models and benchmarks.}
\label{tab:simulation_hyperparameters}
\end{table*}

\pdfanchor{sim-setup}
\subsection{Simulation Experimental Settings}
\label{sim-setup}

Table~\ref{tab:simulation_hyperparameters} summarizes the complete configurations for LingBot-VLA~\cite{wu2026pragmatic}, ABot-M0~\cite{yang2026abot}, and ACT~\cite{zhao2023learning} on RoboTwin 2.0~\cite{chen2025robotwin}, as well as $\pi_{0.5}$~\cite{intelligence2025pi_} on LIBERO~\cite{liu2023libero}. 
On RoboTwin 2.0, BCP is trained under the Clean setting and evaluated under both the Clean and Randomized settings. For LIBERO, BCP is trained on LIBERO and evaluated on both LIBERO and LIBERO-PRO~\cite{zhou2025liberopro}.
The base models use their corresponding action-generation configurations. 
LingBot-VLA, ABot-M0, and ACT employ a prediction horizon of 50 on RoboTwin 2.0, whereas $\pi_{0.5}$ uses a prediction 
horizon of 10 on LIBERO.  
All simulation experiments are implemented on top of RLinf~\cite{yu2026rlinf}, an open-source framework for reinforcement learning of VLA.

-

\begin{figure}[t]
\centering

\begin{minipage}[c]{0.99\linewidth} 
\centering
\includegraphics[width=1.0\textwidth]{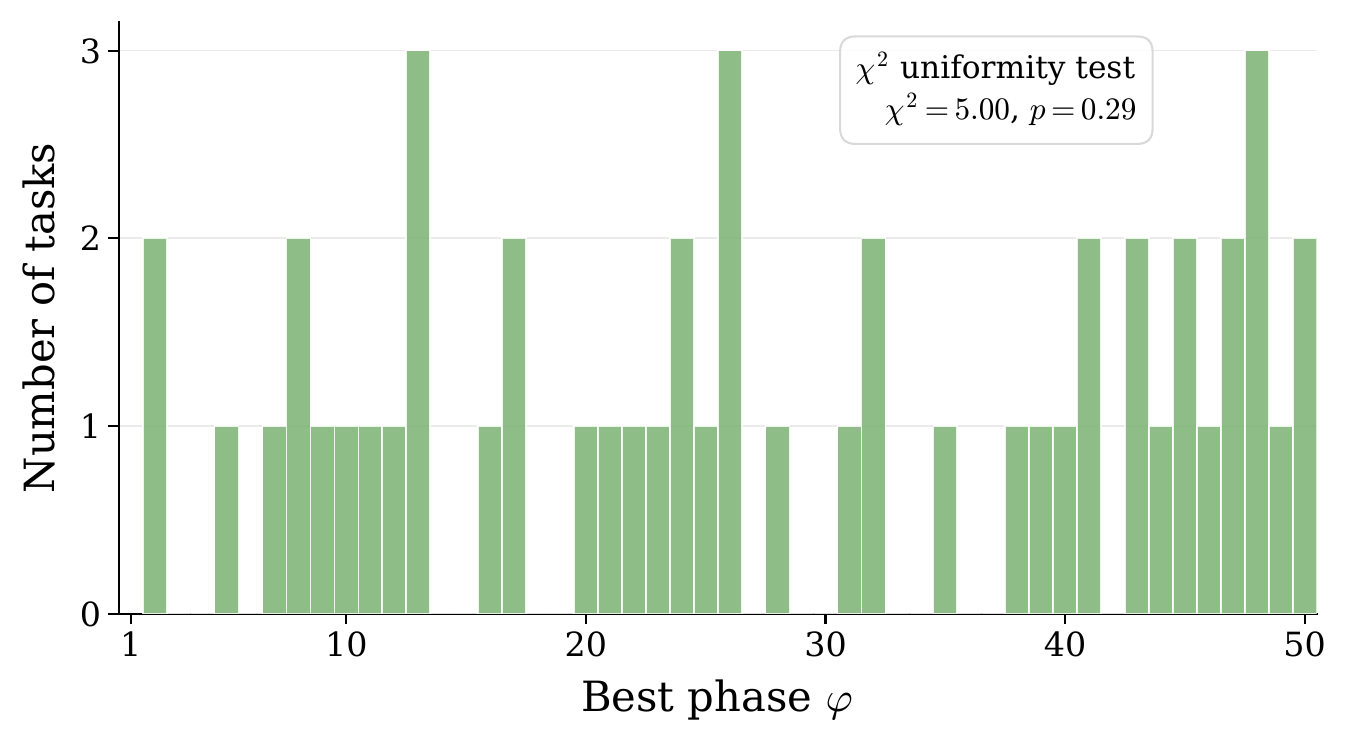}
\end{minipage}

\setlength{\abovecaptionskip}{4pt}
\setlength{\belowcaptionskip}{-16pt}
\caption{Distribution of the best-performing phases across 50 RoboTwin 2.0 tasks. The best phases are broadly distributed. A chi-square goodness-of-fit test does not reject the hypothesis of a uniform distribution ($\chi^2=5.00$, $p=0.287$).}
\label{fig:best_phase_distribution}
\end{figure}

\pdfanchor{additional-results}
\section{Additional Experimental Results}
\label{additional-results}

\pdfanchor{phase-distribution}
\subsection{Distribution of Best Phases in the Phase-Shift Experiment}
\label{phase-distribution}

To further analyze the phase-shift experiment presented in the main paper, we record the best-performing phase for each of the 50 RoboTwin 2.0 tasks by sweeping all phases under the same fixed 50-step execution horizon. As shown in Figure~\ref{fig:best_phase_distribution}, the best phases are broadly distributed across the entire phase range, without any clear concentration or consistent pattern. We further conduct a chi-square goodness-of-fit test against a uniform distribution. The test yields $\chi^2=5.00$ and $p=0.287>0.05$, meaning that we cannot reject the null hypothesis of uniformity. Thus, the optimal phases exhibit no statistically significant preference for any particular region of the phase space.

This result further supports the observation that critical moments occur at different times across tasks and stages. Consequently, no single fixed phase can consistently align replanning boundaries with all precision-critical manipulation stages. This motivates BCP to determine the execution horizon for each action chunk and dynamically adjust the replanning timing of the VLA, allowing the robot to obtain an updated observation before critical manipulation stages.

\pdfanchor{task-results}
\subsection{Detailed Task-Wise Results on RoboTwin 2.0}
\label{task-results}

Table~\ref{tab:task_wise_results} reports the complete task-wise results of LingBot-VLA with and without BCP on all 50 RoboTwin 2.0 tasks.

Under the Clean setting, BCP improves the overall average from 89.88\% to \textbf{93.94\%}. The largest improvement is observed on \emph{Hanging Mug}, where the success rate increases from 41\% to 87\%. Substantial gains are also achieved on \emph{Blocks Ranking Size} (70\% to 88\%), \emph{Turn Switch} (60\% to 72\%), \emph{Click Alarm Clock} (80\% to 91\%), and \emph{Place Object Scale} (80\% to 91\%). These tasks involve precision-sensitive grasping, placement, or interaction stages, for which aligning replanning with task progress is particularly important.

BCP also exhibits consistent improvements under the Randomized setting, despite being trained only under the Clean setting. It raises the average success rate from 88.78\% to 92.84\%. Notable improvements include \emph{Hanging Mug} (36\% to 62\%), \emph{Blocks Ranking Size} (71\% to 89\%), \emph{Beat Block Hammer} (78\% to 92\%), and \emph{Place A2B Left} (67\% to 81\%).
These results indicate that BCP does not merely
overfit to the specific visual conditions seen during training.
Instead, the improvement suggests that the policy captures
task-stage-level knowledge about when a predicted action chunk should be trusted and when replanning 
is necessary.

\begin{table}[t]
    \centering
    \setlength{\abovecaptionskip}{4pt}
    \small
    \setlength{\tabcolsep}{2pt}
    \begin{tabular}{lcccc}
        \toprule
        Method & \makecell{VLA \\ Calls} $\downarrow$ & \makecell{Exec. \\ Steps} $\downarrow$ & Runtime (s) $\downarrow$ & \makecell{Success \\ Rate} $\uparrow$  \\
        \midrule
        SFT       &13.525  &676.250 & 26.21 & 41\%  \\
        \midrule
        RL (Fixed-horizon)    &9.392 & 469.583 & 18.20 & 79\%  \\
        \midrule
        RL (Softmax Head)  & 12.142& 464.708 & 20.71 & 78\%   \\
        RL (BC Head)    &13.825 &427.750 & 21.55 & 83\%  \\
        RL (BC Head) + RER      &10.158 &430.958 & 18.17 & 87\%  \\
        \bottomrule
    \end{tabular}
    \caption{Runtime efficiency analysis on the \emph{Hanging Mug} task under the Clean setting of RoboTwin 2.0.}
    \label{tab:ablation_runtime}
\end{table}

\pdfanchor{runtime-analysis}
\subsection{Runtime Efficiency Analysis}
\label{runtime-analysis}

Table~\ref{tab:ablation_runtime} extends the ablation study in the main paper by reporting the average number of VLA calls, executed steps, and overall runtime for each variant on the \emph{Hanging Mug} task. The results reveal that the different variants improve task performance through distinct mechanisms and consequently exhibit different efficiency characteristics.

The SFT policy achieves the lowest success rate of 41\%. Its frequent failures often lead to prolonged unsuccessful exploration before the episode terminates, causing the policy to accumulate the largest numbers of VLA calls and executed control steps. Consequently, it incurs the highest overall runtime of 26.21\,s. RL (Fixed-horizon) directly updates the action expert, allowing the base policy to generate more effective actions and substantially improving the success rate. However, it retains the fixed 50-step execution horizon, leading to more executed steps than the adaptive-horizon variants.

RL (Softmax Head) and RL (BC Head), in contrast, keep the action-generation policy fixed and improve execution by adaptively adjusting the execution horizon. Both variants can shorten a chunk so that the VLA replans before critical manipulation stages. Nevertheless, when trained only with task-level success, both policies exhibit a conservative tendency toward short horizons. Although frequent replanning reduces reliance on previously predicted actions, it also increases the number of expensive VLA queries. Consequently, the Softmax Head and BC Head require longer runtimes despite executing fewer control steps than the fixed-horizon variant.

The complete RL (BC Head) + RER variant addresses this efficiency issue by incorporating the proposed Replanning-Efficiency Reward. RER explicitly penalizes inefficient short-horizon decisions and discourages unnecessary VLA queries, while its efficiency-aware reward provides more informative advantage signals than binary task success alone. This reduces the average number of VLA calls from 13.825 to 10.158 while maintaining a similar number of executed control steps. Consequently, the complete method achieves both the highest success rate of 87\% and the lowest overall runtime of 18.17\,s.

\pdfanchor{hyperparameter-study}
\subsection{Hyperparameter}
\label{hyperparameter-study}

Table~\ref{tab:hyperparameter_study} studies the influence of the BCP architecture, candidate execution horizons, rollout temperature, and LR schedule on the \emph{Hanging Mug} task under the Clean setting of RoboTwin 2.0. All variants use the same RER configuration, with $\delta_{+}=0.7$ and $\delta_{-}=0.3$, and performance is insensitive to moderate variations in these two hyperparameters.

For the BCP architecture, the two-layer Transformer achieves the highest success rate of 87\%. Reducing the encoder to one layer lowers the success rate to 82\%, suggesting that a shallow model is less effective at integrating visual-language context with action-level features. Increasing the depth to four or six layers also reduces performance to 84\% and 83\%, respectively. This indicates that additional model depth is unnecessary for execution-horizon selection and may make optimization more difficult. Overall, two Transformer layers provide an effective balance between representation capacity and training stability.

The choice of candidate horizons also has a substantial effect. Replacing the fine-grained candidate set $\{15,20,25,\ldots,50\}$ with the coarser set $\{20,30,40,50\}$ reduces the success rate from 87\% to 78\%. The finer spacing provides BCP with greater flexibility to place replanning boundaries immediately before precision-critical manipulation stages, whereas coarse candidates may force the policy to stop either too early or too late.

We further examine the RL optimization settings. Increasing the rollout temperature from 1.0 to 1.6 reduces the success rate from 87\% to 82\%. Although a higher temperature encourages broader exploration, excessive randomness in horizon selection can produce less reliable trajectories and weaken the learning signal. Meanwhile, replacing the constant learning-rate schedule with a cosine schedule results in an 82\% success rate. 

Based on these results, we use a two-layer Transformer, the candidate horizons $\{15,20,25,\ldots,50\}$, a rollout temperature of 1.0, and a constant learning-rate schedule for all RoboTwin 2.0 tasks.

\begin{table}[t]
    \centering
    \setlength{\abovecaptionskip}{4pt}
    \setlength{\belowcaptionskip}{-15pt}
    \small
    \setlength{\tabcolsep}{1pt}
    \begin{tabular}{ccccc}
        \toprule
        \makecell{Transformer \\ Layers in BCP}
        & \makecell{Candidate \\ Horizons}
        & \makecell{Rollout \\ Temperature}
        & \makecell{LR \\ Schedule}
        & SR \\
        \midrule

        1
        & \{15, 20, 25,..., 50\}
        & 1.0
        & Constant
        & 82\% \\

        \textbf{2}
        & $\bm{\{15, 20, 25,..., 50\}}$
        & \textbf{1.0}
        & \textbf{Constant}
        & \textbf{87\%} \\

        4
        & \{15, 20, 25,..., 50\}
        & 1.0
        & Constant
        & 84\% \\

        6
        & \{15, 20, 25,..., 50\}
        & 1.0
        & Constant
        & 83\% \\

        \midrule

        2
        & $\{20,30,40,50\}$
        & 1.0
        & Constant
        & 78\% \\

        2
        & \{15, 20, 25,..., 50\}
        & 1.0
        & Cosine
        & 82\% \\

        2
        & \{15, 20, 25,..., 50\}
        & 1.6
        & Constant
        & 82\% \\

        \bottomrule
    \end{tabular}
    \caption{
        Hyperparameter study on the \emph{Hanging Mug} task under the Clean setting
        of RoboTwin 2.0. All experiments use the same configuration of our RER.
    }
    \label{tab:hyperparameter_study}
\end{table}

\pdfanchor{limitations}
\section{Limitations and Future Work}
\label{limitations}

Our BCP adjusts when the base VLA should stop executing the current action chunk and replan, but it does not modify the actions generated by the VLA. Its effectiveness therefore relies on the base VLA producing action chunks that are reasonable and contain useful actions for completing the task. When the predicted actions are fundamentally incorrect, selecting a different execution horizon or triggering replanning at a more appropriate time may still be insufficient to correct the trajectory. Thus, BCP primarily addresses failures caused by the misalignment between replanning timing and critical manipulation stages, rather than errors originating from the action-generation capability of the base VLA.

An interesting direction is to extend adaptive execution-horizon learning to World Action Models~\cite{bi2025motus,li2026causal,kim2026cosmos,ye2026world}. These models also generate action chunks for multi-step execution and therefore face a similar question of how long a predicted chunk should be executed before obtaining a new observation and replanning. Extending BCP to this setting would provide a principled mechanism for dynamically aligning replanning boundaries with critical task stages.

\begin{table*}[t]
    \centering
    \normalsize
    \setlength{\tabcolsep}{15pt}
    \renewcommand{\arraystretch}{0.92}
    \begin{tabular}{lcccc}
        \toprule
        \multirow{2}{*}{\textbf{Simulation Task}} &
\multicolumn{2}{c}{LingBot-VLA} &
\multicolumn{2}{c}{\textbf{LingBot-VLA}$^{\dagger}$} \\
\cmidrule(lr){2-3}
\cmidrule(lr){4-5}
& Clean & Rand. & Clean & Rand. \\
        \midrule
        \textit{Adjust Bottle}              & 96\%    & 97\%    & 97\%    & 97\%    \\
        \textit{Beat Block Hammer}           & 82\%    & 78\%    & 91\%    & 92\%    \\
        \textit{Blocks Ranking RGB}          & 88\%    & 88\%    & 98\%    & 98\%    \\
        \textit{Blocks Ranking Size}         & 70\%    & 71\%    & 88\%    & 89\%    \\
        \textit{Click Alarm Clock}           & 80\%    & 81\%    & 91\%    & 82\%    \\
        \textit{Click Bell}                  & 99\%    & 100\%   & 100\%   & 100\%   \\
        \textit{Dump Bin BigBin}            & 98\%    & 98\%    & 97\%    & 98\%    \\
        \textit{Grab Roller}                 & 100\%   & 100\%   & 100\%   & 100\%   \\
        \textit{Handover Block}              & 90\%    & 92\%    & 98\%    & 93\%    \\
        \textit{Handover Mic}                & 94\%    & 87\%    & 100\%   & 100\%   \\
        \textit{Hanging Mug}                 & 41\%    & 36\%    & 87\%    & 62\%    \\
        \textit{Lift Pot}                    & 100\%   & 99\%    & 100\%   & 100\%   \\
        \textit{Move Can Pot}                & 90\%    & 91\%    & 90\%    & 93\%    \\
        \textit{Move Pillbottle Pad}        & 98\%    & 99\%    & 97\%    & 98\%    \\
        \textit{Move Playingcard Away}      & 98\%    & 98\%    & 99\%    & 98\%    \\
        \textit{Move Stapler Pad}            & 75\%    & 70\%    & 82\%    & 78\%    \\
        \textit{Open Laptop}                 & 98\%    & 96\%    & 98\%    & 98\%    \\
        \textit{Open Microwave}              & 93\%    & 88\%    & 93\%    & 96\%    \\
        \textit{Pick Diverse Bottles}        & 90\%    & 89\%    & 91\%    & 91\%    \\
        \textit{Pick Dual Bottles}           & 100\%   & 99\%    & 100\%   & 99\%    \\
        \textit{Place A2B Left}              & 69\%    & 67\%    & 76\%    & 81\%    \\
        \textit{Place A2B Right}             & 81\%    & 73\%    & 79\%    & 84\%    \\
        \textit{Place Bread Basket}          & 89\%    & 90\%    & 98\%    & 95\%    \\
        \textit{Place Bread Skillet}         & 90\%    & 83\%    & 92\%    & 89\%    \\
        \textit{Place Burger Fries}          & 99\%    & 98\%    & 100\%   & 100\%   \\
        \textit{Place Can Basket}            & 81\%    & 83\%    & 84\%    & 81\%    \\
        \textit{Place Cans Plastic Box}      & 98\%    & 100\%   & 100\%   & 99\%    \\
        \textit{Place Container Plate}       & 92\%    & 91\%    & 99\%    & 96\%    \\
        \textit{Place Dual Shoes}            & 81\%    & 85\%    & 90\%    & 88\%    \\
        \textit{Place Empty Cup}             & 100\%   & 98\%    & 100\%   & 100\%   \\
        \textit{Place Fan}                   & 97\%    & 96\%    & 98\%    & 98\%    \\
        \textit{Place Mouse Pad}             & 86\%    & 81\%    & 92\%    & 88\%    \\
        \textit{Place Object Basket}         & 91\%    & 87\%    & 93\%    & 87\%    \\
        \textit{Place Object Scale}          & 80\%    & 83\%    & 91\%    & 89\%    \\
        \textit{Place Object Stand}          & 94\%    & 98\%    & 97\%    & 98\%    \\
        \textit{Place Phone Stand}           & 94\%    & 89\%    & 95\%    & 92\%    \\
        \textit{Place Shoe}                  & 97\%    & 99\%    & 96\%    & 98\%    \\
        \textit{Press Stapler}               & 90\%    & 85\%    & 94\%    & 88\%    \\
        \textit{Put Bottles Dustbin}         & 96\%    & 93\%    & 96\%    & 95\%    \\
        \textit{Put Object Cabinet}          & 93\%    & 92\%    & 93\%    & 94\%    \\
        \textit{Rotate Qrcode}              & 93\%    & 93\%    & 94\%    & 96\%    \\
        \textit{Scan Object}                 & 92\%    & 94\%    & 95\%    & 95\%    \\
        \textit{Shake Bottle Horizontally}   & 100\%   & 100\%   & 100\%   & 100\%   \\
        \textit{Shake Bottle}                & 100\%   & 99\%    & 100\%   & 99\%    \\
        \textit{Stack Blocks Three}          & 92\%    & 94\%    & 96\%    & 97\%    \\
        \textit{Stack Blocks Two}            & 100\%   & 100\%   & 99\%    & 100\%   \\
        \textit{Stack Bowls Three}           & 89\%    & 84\%    & 87\%    & 86\%    \\
        \textit{Stack Bowls Two}             & 98\%    & 99\%    & 98\%    & 98\%    \\
        \textit{Stamp Seal}                  & 89\%    & 88\%    & 96\%    & 96\%    \\
        \textit{Turn Switch}                 & 60\%    & 60\%    & 72\%    & 73\%    \\
        \midrule
        \textbf{Average}
        & \textbf{89.88\%}
        & \textbf{88.78\%}
        & \makecell[t]{\textbf{93.94\%} \\ \textbf{\textcolor{improvegreen}{(+4.06\%)}}} 
        & \makecell[t]{\textbf{92.84\%} \\ \textbf{\textcolor{improvegreen}{(+4.06\%)}}} \\
        \bottomrule
    \end{tabular}
    \caption{Success rate comparison on RoboTwin 2.0. ${\dagger}$ denotes the addition of BCP.}
    \label{tab:task_wise_results}
\end{table*}

\bibliography{2027}